\documentclass[runningheads]{llncs}
\usepackage{graphicx}
\usepackage{enumitem}
\usepackage{xcolor}
\usepackage{appendix}
\usepackage{bm}
\usepackage{amsmath}
\usepackage{booktabs}
\usepackage{tabularx}
\usepackage{makecell}
\usepackage{mathtools}
\usepackage{tcolorbox}
\usepackage{hyperref}
\usepackage{xurl}
\newcolumntype{P}[1]{>{\centering\arraybackslash}p{#1}}

\newcommand{\codebold}[1]{{\ttfamily\bfseries #1}}
\newcommand{\shrink}{\vspace{-2mm}}

\newcommand{\bn}{\codebold{BN}}
\newcommand{\bnplus}{\codebold{BN\textsuperscript{++}}}
\newcommand{\ff}{\codebold{FF-{\allowbreak}discr-{\allowbreak}text}}
\newcommand{\bngen}{\codebold{BN-{\allowbreak}gen-{\allowbreak}text}}
\newcommand{\bndiscr}{\codebold{BN-{\allowbreak}discr-{\allowbreak}text}}
\newcommand{\bngenabl}{\codebold{BN-{\allowbreak}gen-{\allowbreak}text\textsuperscript{$-$}}}
\newcommand{\bndiscrabl}{\codebold{BN-{\allowbreak}discr-{\allowbreak}text\textsuperscript{$-$}}}

\begin{document}
\title{Clinical Reasoning over Tabular Data and Text with Bayesian Networks}
\titlerunning{Clinical reasoning over Tabular Data \& Text w. BNs}
%
\author{Paloma Rabaey\inst{1}$^{,}$*\orcidID{0000-0001-6064-0788} \and
Johannes Deleu\inst{1} \and
Stefan Heytens\inst{2}\orcidID{0000-0003-1097-4987} \and
Thomas Demeester\inst{1}\orcidID{0000-0002-9901-5768}}
\authorrunning{P. Rabaey et al.}
%
\institute{IDLab, Dept. of Information Technology, Ghent University - imec, Ghent, Belgium \\ \and 
Dept. of Public Health and Primary Care, Ghent University, Ghent, Belgium \\
* Corresponding author,
\email{\{first.last\}@ugent.be}}
\maketitle              
\begin{abstract}
\shrink \shrink \shrink \shrink
Bayesian networks are well-suited for clinical reasoning on tabular data, but are less compatible with natural language data, for which neural networks provide a successful framework.
This paper compares and discusses strategies to augment Bayesian networks with neural text representations, both in a generative and discriminative manner.
This is illustrated with simulation results 
for a primary care use case (diagnosis of pneumonia) and discussed 
in a broader clinical context.
\shrink
\keywords{Clinical reasoning  \and Bayesian networks \and Neural networks \and Text representations.}
\shrink\shrink
\end{abstract}
\section{Introduction}
\shrink 

The process of clinical reasoning lies at the heart of many interactions between a clinician and their patient \cite{clinical_reasoning_yazdani}. Clinical reasoning is the process by which a clinician integrates their own knowledge with patient information (like symptoms, objective medical evidence, background, medical history...), to arrive at a diagnosis and subsequent therapeutic options \cite{clinical_reasoning}. 
Cognitive biases and knowledge deficits can cause errors in clinical reasoning, causing the clinician to arrive at an incorrect diagnosis \cite{clinical_reasoning_errors}. To help clinicians avoid these pitfalls, it can help to (partially) automate the process of clinical reasoning \cite{clinical_judgement,computer_aided_diagnosis}. Bayesian networks (BNs) are ideally suited for this task, given 
(i) their ability to model complex problems involving uncertainty, (ii) their ability to combine data and expert knowledge, and (iii) their interpretable graphical structure \cite{kyrimi_scoping_BN}. 
However, a key factor limiting the adoption of BNs in clinical practice is their inadequacy to deal with realistic medical data \cite{BNadoption}, 
often a mix between structured tabular variables (disease codes, timestamps, demographic features, lab results...) and unstructured text (consultation notes, discharge summaries...) \cite{extracting_information_EMR}. 
Encoding the information contained in the unstructured text into structured variables not only requires (considerable) effort, but also inevitably results in loss of information. 



In this work, we explore how to integrate unstructured text data in Bayesian networks, to facilitate joint clinical reasoning over structured tabular data and unstructured text. To this end, we investigate a relevant use case in primary care: diagnosis of pneumonia. We create an artificial yet realistic dataset, allowing us to control several aspects of the data generation process. 
This allows us to investigate the impact of different modeling 
approaches to integrate text in the clinical reasoning process, and discuss their advantages and pitfalls. 
By keeping the use case highly tangible for a clinical audience, we aim to lower the bar toward real-world medical applications of the presented technology.

Our main contribution is the study of different approaches to integrate the neural representation of a textual variable in the BN. In particular, we compare the properties of adding the text with a generative model (in the space of neural text representations, 
fitted alongside the BN) vs.~a discriminative model (a text classifier jointly trained with the BN). We evaluate the performance of both approaches on the prediction of pneumonia in a toy setting, and compare with baselines which are either missing the text component or the BN structure. Based on the presented results, we discuss (i) the advantages of including unstructured text, (ii) the properties of different approaches to achieve this, and (iii) the overall idea of performing Bayesian inference for automated clinical reasoning involving textual data.

\shrink
\section{Related Work} \label{sec:related_work}
\shrink

Since the topic of this paper touches on multiple different research domains, we position our work in regards to the most relevant domains, without providing an exhaustive overview of all related research.

\textbf{Clinical reasoning:} \quad 
This work follows the interpretation of clinical reasoning as an analytical process, where a clinician weighs up every piece of evidence to reject or confirm a certain diagnostic hypothesis \cite{evidence_based_medicine,clinical_reasoning_GP}. 
Starting from a set of differential diagnoses, each with their own prior probability reflecting their prevalence in the population, clinical reasoning comes down to updating
the likelihood of each diagnosis with every new piece of evidence that comes in, using Bayes' rule. 
This results in 
a posterior likelihood for each diagnosis, which the clinician takes into account for planning further steps. 


\textbf{Bayesian networks:} \quad 
BNs form the perfect tool to formalize the process outlined above \cite{kyrimi_scoping_BN}. Their interpretable graph structure can help keep track of independencies between certain types of evidence and particular diagnoses, and inference in BNs follows Bayes' rule.
BNs have been used to model a wide range of medical conditions in research settings  \cite{BN_healthcare_condition}, including respiratory diseases such as pneumonia and Covid-19 \cite{BN_respiratory}. However, their deployment for clinical decision making in practice remains limited, partly due to real-world data challenges 
\cite{BNadoption}. 

\textbf{Clinical unstructured text:}\quad The last few decades have seen an abundance of electronic medical records being collected in clinical practice, which form a useful source of data 
to build clinical decision support systems (CDSS). These records are usually made up of structured data (disease codes, dates, treatment codes...), as well as 
free text \cite{extracting_information_EMR}. Studies have shown that ignoring the information present in free text records can results in data loss and bias in CDSS \cite{omission_free_text}. Nevertheless, a large majority of CDSS either completely disregards this unstructured text \cite{CDS_infectious_diseases} or applies information extraction techniques to turn the text into tabular format, which then serves as input to the CDSS 
\cite{extracting_information_EMR}. 
Turning unstructured text into structured variables using information extraction methods (see, e.g., \cite{sterckx2020_clinical_IE}) and then building a predictive model on top of the structured features has been applied to learning clinical BNs in the past \cite{influenza_detection,health_knowledge_graph}. 
Our work focuses on integrating the full unstructured text, removing the need for this information extraction 
step. Some CDSS are built on raw unstructured text, yet they often fail to integrate it with the structured portion of the data \cite{CDS_infectious_diseases,clinical_text_classification}. Zhang et al.~manage to successfully integrate both through a multi-modal recurrent neural network that combines embeddings of clinical text with static and time-varying features in the electronic medical record, outputting a full patient representation that can be used for further downstream prediction \cite{combining_structured_unstructured}. We also represent clinical text through neural representations, though we use a BN in combination with feed-forward neural components to integrate these text embeddings with the static tabular features in the reasoning process. 


\textbf{Neuro-Symbolic AI:}\quad The integration of reasoning and learning has seen considerable progress in recent years, in the field of Neuro-Symbolic (NeSy) AI \cite{Marra2024_NeSyAI}. One strongly related contribution is the DeepProbLog (DPL) framework \cite{Manhaeve2019_DPL}. The authors show how a probabilistic logic program can be extended with neural predicates, whereby a neural network converts an unstructured data item (like an image) into class probabilities, that are then treated as regular predicates in the logic program. Importantly, the parameters of the logic program and of the neural networks encoding unstructured data can be jointly trained. The discriminative model for integrating text nodes into BNs in this paper corresponds to the neural predicates approach, since a BN can be seen as a special case of a probabilistic program. In contrast, we also compare this approach with a generative model in text representation space, and provide a targeted discussion from the perspective of clinical reasoning.




\shrink
\section{Use Case and Data Description} \label{sec:data_gen}
\shrink

\begin{figure}
    \includegraphics[width=\textwidth]{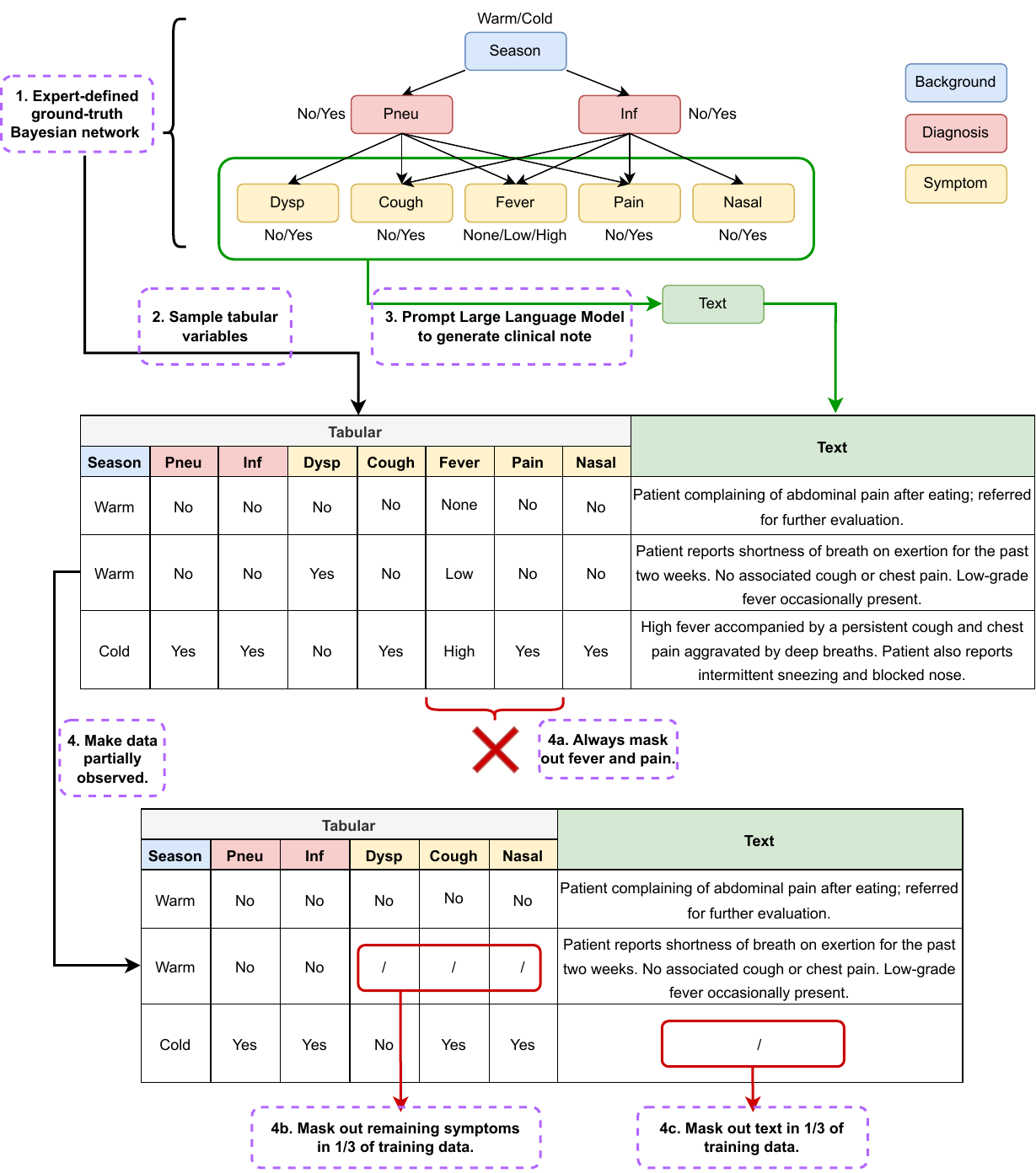}
    \caption{Key steps in generating the artificial dataset, where each sample consists of both tabular variables and corresponding clinical text descriptions. With help of an expert, we define a Bayesian network (BN) simulating the pneumonia use case (\textbf{step 1}). We sample the tabular variables (background, diagnoses and symptoms) from the distribution defined by this BN (\textbf{step 2}), prompting a large language model (GPT3.5 \cite{instructGPT}) to generate realistic but fictitious consultation notes based on the sampled symptoms (\textbf{step 3}). We repeat steps 1 to 3 to generate 4000 training samples and 1000 test samples. Finally, to induce property (ii) of realistic medical data (see Section \ref{sec:data_gen}), we remove two symptoms, $fever$ and $pain$, from the tabular portion of the data, ensuring they are never encoded and only observed through the text (\textbf{step 4a}). From now on, when we talk about symptoms, we take this to mean the symptoms $dysp$, $cough$, $nasal$, unless explicitly stated otherwise. 
    For the training set only, we partially mask out the remaining symptoms (\textbf{step 4b}) and the text (\textbf{step 4c}) in a complementary subset of the training samples. Each sample now represents a fictional patient encounter, consisting of one background feature ($season$), two diagnoses ($pneu$ and $inf$), three symptoms ($dysp$, $cough$ and $nasal$, partially unobserved) and a textual description ($text$, partially unobserved). The text contains additional context on the three encoded symptoms, as well as describing two additional unencoded symptoms $fever$ and $pain$.} 
    \label{fig:data_generation}
\end{figure}

Our use case focuses on diagnostic clinical reasoning performed by a general practitioner (GP) \cite{clinical_reasoning_GP}. 
One non-trivial task in primary care is distinguishing pneumonia from an upper respiratory tract infection (also known as the common cold), where the former is more serious and calls for treatment with antibiotics. When a patient presents with respiratory symptoms, a GP will apply clinical reasoning based on these symptoms and a short clinical examination, ordering the necessary additional testing or starting a treatment only if the probability for pneumonia exceeds a certain threshold. 


We create an artificial dataset that allows us to study automation of the clinical reasoning process for the pneumonia use case, in the presence of unstructured text resembling consultation notes taken by the GP during a patient encounter. Figure \ref{fig:data_generation} shows the data generation process. Its caption describes the four key steps, and a more detailed explanation is given in Appendices \ref{sec:app:data_gen} and \ref{sec:app:data_gen_prompt}.
We aim to mirror the following properties of realistic medical data: 
(i) the data contains 
structured tabular variables and/or 
unstructured text, (ii) information in the text is only partially encoded in the structured variables, and
(iii) the text contains additional context on the patient's background and symptoms, complementing the information encoded in the tabular variables. The final train and test datasets are available in our Github repository: \url{https://github.com/prabaey/bn-text}.

\shrink
\section{Augmenting BNs with Text Representations} \label{sec:models}
\shrink

\begin{figure}[t]
    \includegraphics[width=\textwidth]{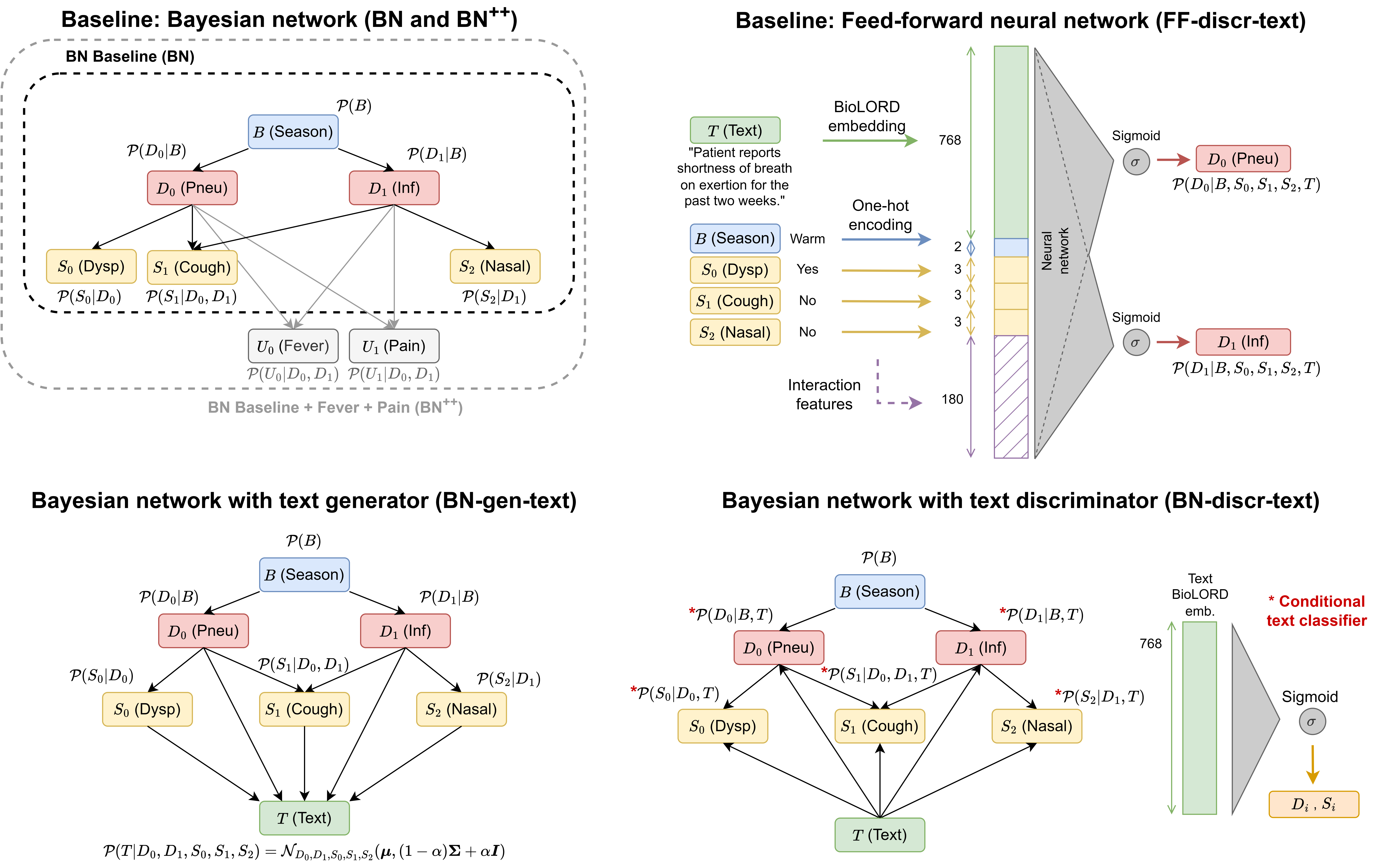}
    \caption{Schematic depiction of all models. The top row presents our baselines \bn{}, \bnplus{} and \ff{}. The bottom row shows \bngen{} and \bndiscr{}, two variants of a Bayesian network augmented with text representations. 
    \shrink\shrink
    }
    \label{fig:models}
\end{figure}

We propose two model architectures that are able to integrate text in a Bayesian network: \bngen{} (Section \ref{sec:model_gen}) and \bndiscr{} (Section \ref{sec:model_discr}). 
Both models incorporate text through a single-vector text embedding, either modeling its distribution directly or learning classifiers with these representations as an input.
As shown in Figure \ref{fig:models}, we compare them with three baseline models. Our first baseline \bn{} is a standard Bayesian network without text variables, trained only on 
the partially observed tabular features. Its extension \bnplus{} is trained on a version of the training set where the symptoms $fever$ and $pain$ are exceptionally \textit{not} masked out, forming an upper bound to the performance of all other models, which never get to directly observe these two symptoms. The last baseline \ff{} is a discriminative feed-forward neural network which takes both tabular features (one-hot encoded) and text (as a \textit{BioLORD} embedding) as an input, and outputs a prediction for $pneu$ or $inf$. Details on the baseline models \bn{}, \bnplus{} and \ff{} can be found in Appendix \ref{sec:model_BN} and \ref{sec:model_FF}.

All models are trained on the final dataset shown in Figure \ref{fig:data_generation} (with only the symptoms $dysp$, $cough$ and $nasal$, partially observed) except for \bnplus{} (where $fever$ and $pain$ are added, as described above). 
During inference, each model computes a posterior distribution for each diagnosis given some set of evidence. 
For readability, we 
represent the diagnoses by $D_i$ ($i\in\{0, 1\}$) with $D_0$ ($pneu$) and $D_1$ ($inf$), symptoms as $S_0$ ($dysp$), $S_1$ ($cough$) and $S_2$ ($nasal$), background as $B$ ($season$) and text as $T$. 
We discuss how each model is able to calculate the following posterior diagnostic probabilities: 
\shrink
\begin{itemize}
    \item $\mathcal{P}(D_i\mid B,S_0,S_1,S_2)$: take only background and symptoms as evidence.
    \item $\mathcal{P}(D_i\mid B,S_0,S_1,S_2,T)$: take background, symptoms and text as evidence.
    \item $\mathcal{P}(D_i\mid B,T)$: take background and text as evidence.
    \shrink
\end{itemize}

\shrink
\subsection{Bayesian Network with Text Generator (\bngen{})} \label{sec:model_gen}

\textbf{Training:} \quad 
In the \bngen{} model, a text node is added to the \bn{} baseline, 
conditioned on all diagnoses and symptoms. 
The conditional distributions for all tabular variables are trained using Maximum Likelihood Estimation, 
as a standard Bayesian network (see Appendix \ref{sec:model_BN}). To obtain a vector representation for the text, we use \textit{BioLORD}, which is a pre-trained language model that produces semantic single-vector representations for clinical sentences and biomedical concepts \cite{BioLORD}. 32 separate multivariate Gaussians, one for each possible combination of the values for the two diagnoses and three symptoms, are fitted to the text embeddings to obtain the distribution $\mathcal{P}(T\mid D_0,D_1,S_0,S_1,S_2)$. This basic model allows us to get the probability density of unseen text embeddings and even sample new ones, although those cannot be directly decoded into text. 
To learn each Gaussian, we select all samples in the training set that match a particular condition and fit the mean $\mu$ and covariance matrix $\bm{\Sigma}$ to the corresponding text embeddings. 
The estimated covariance matrix $\Sigma$ is regularized as follows
%
%
\vspace{-2mm}
\begin{equation} \label{eq:gauss_gen}
\mathcal{P}(T\mid D_0,D_1,S_0,S_1,S_2) = \mathcal{N}_{D_0,D_1,S_0,S_1,S_2}(\bm{\mu}, (1-\alpha)\bm{\Sigma} + \alpha \bm{I})
\vspace{-2mm}
\end{equation}
where the hyperparameter $\alpha$ allows tuning the contribution of the individual variances of the text representation dimensions.

\noindent
\textbf{Inference:} \quad After training, we can calculate the posterior for either diagnosis $D_i$ given a set of evidence by applying Bayes' rule and marginalizing over the learned joint distribution. For $\mathcal{P}(D_i\mid B,S_0,S_1,S_2)$ the calculation is the same as in a standard BN, since the conditional text distribution $\mathcal{P}(T\mid D_0,D_1,S_0,S_1,S_2)$ is integrated out. $\mathcal{P}(D_i\mid B,S_0,S_1,S_2,T)$ and $\mathcal{P}(D_i\mid B,T)$ on the other hand do evaluate the conditional text distribution. The detailed equations for all posteriors can be found in Appendix \ref{sec:app:gen}. 

\shrink
\subsection{Bayesian Network with Text Discriminator (\bndiscr{})} \label{sec:model_discr}

\textbf{Training:} \quad In the \bndiscr{} model, we augment the \bn{} baseline by conditioning all diagnoses and symptoms on the text embedding. This contrasts with the \bngen{} approach, where we augment the \bn{} baseline with distributions of text embeddings conditioned on the diagnoses and symptoms. While this arc reversal renders the two BNs semantically non-equivalent, all independence relations between the non-text nodes remain intact.
Each of the conditional distributions is modeled as a set of discriminative neural text classifiers, one for each configuration of the tabular parent variables, meaning there are 12 in total. For example, we model $\mathcal{P}(D_0\mid B=warm,T)$ and $\mathcal{P}(D_0\mid B=cold,T)$ as two separate feed-forward neural networks that take the \textit{BioLORD} text embedding as an input, and learn to predict the diagnosis probability for $D_0$ at the output. All parameters are optimized jointly by maximizing the likelihood $\mathcal{P}(B,D_0,D_1,S_0,S_1,S_2\mid T)$ (see Appendix \ref{sec:app:discr}) based on the training data. By making this likelihood conditional on text, we refrain from having to learn a prior distribution $\mathcal{P}(T)$ of the text embeddings. 

\noindent
\textbf{Inference:} \quad $\mathcal{P}(D_i\mid B,S_0,S_1,S_2,T)$ is again obtained by applying Bayes' rule and marginalizing over the joint distribution (conditional on text). The trained classifiers are used to evaluate the probabilities needed during inference. Strictly speaking, conditioning on the text node means that $\mathcal{P}(D_i\mid B,S_0,S_1,S_2)$ cannot be computed. 
We circumvent this issue by conditioning on the embedding of the empty text ``'' in case no text is observed. The classifiers learn to take this into account, 
since an empty text occurs in 1/3 of the training data. 
Finally, $\mathcal{P}(D_i\mid B,T)$ is simply the output of one of the two diagnosis classifiers. As before, the detailed equations for all posteriors can be found in Appendix \ref{sec:app:discr}. 

\shrink
\section{Empirical Results and Analysis} \label{sec:results}
\shrink

\textbf{Evaluation:}\quad There are various ways to measure the models' ability of estimating diagnostic probabilities, given the observed background, symptoms and/or textual inputs. This section assumes the real-world scenario with a known (binary) diagnosis on the test set, but no knowledge of ground truth conditional probabilities. We therefore rank all patients in the test set according to the estimated probability of the considered diagnosis, and measure the area under the precision-recall curve for that ranking by comparing with the binary ground truth, i.e., we report the average precision. Results are averaged over 5 training runs with different model initializations. The full code is available in our Github repository: \url{https://github.com/prabaey/bn-text}.

\noindent
\textbf{Results:}\quad 
Table~\ref{tab:res_pneu} presents average precision results for the diagnosis of pneumonia ($D_0$). Corresponding results for the prediction of upper respiratory tract infection ($D_1$) are shown in Table \ref{tab:res_inf} in Appendix \ref{sec:app:res}. Ablation results in terms of connectivity in the network are provided and discussed in Appendix \ref{sec:res_abl}, and details on training and hyperparameter optimization are given in Appendix \ref{sec:app:training}.



\noindent
\textbf{Analysis:}\quad
Comparing $\mathcal{P}(D_0\mid B,S_0,S_1,S_2,T)$ and $\mathcal{P}(D_0\mid B,S_0,S_1,S_2)$ in Table \ref{tab:res_pneu}, we note that both \bngen{} and \bndiscr{} improve over the baseline \bn{}. 
This improvement is thanks to the incorporation of text, which contains information on the symptoms $fever$ and $pain$ that is otherwise never explicitly encoded in the tabular features, yet very useful for diagnosing $pneu$. Indeed, when \bndiscr{} takes both symptoms and text into account, in $\mathcal{P}(D_0\mid B,S_0,S_1,S_2,T)$, its average precision comes close to the upper bound set by the baseline \bnplus{}, with direct access to all 5 encoded symptoms. 
Furthermore, the ablation study in Appendix \ref{sec:res_abl} shows a dramatic performance drop for both \bndiscr{} and \bngen{} when omitting 
the direct relation between diagnoses and text in the network, rendering the model unable to incorporate any complementary text-only information (i.e., on 
$fever$ and $pain$) during inference.


Though the \bngen{} model performs better than the baseline \bn{}, we see two reasons for why it is not on par with \bndiscr{}. 
First, the distribution $\mathcal{P}(T\mid D_0,D_1,S_0,S_1,S_2)$ is made up of 32 conditional Gaussians, each trained on a different subset of text embeddings that occur with a particular (possibly rare) combination of symptom and diagnosis values. 
The \bndiscr{} model has a more modular architecture and does not suffer as much from limited relevant training samples to fit each of its classifiers. Second, a multivariate normal distribution is not the best fit for the text embeddings. 
This is probably also why we see a bigger performance gap between the prediction that incorporates only text ($P(D_0\mid B,T)$) and the one that incorporates both text and symptoms ($P(D_0\mid B,S_0,S_1,S_2,T)$) for the \bngen{} model. 

Interestingly, the \ff{} baseline performs worse than \bndiscr{}.
While the two $pneu$ classifiers in \bndiscr{} can focus on modeling the text given one particular value of the background variable, the \ff{} classifier needs to deal with various configurations of the background and symptoms, missing or present, as well as their interactions with the text, in a single model. This is why \bndiscr{} already improves over \ff{}, even when it only incorporates text during inference. When \bndiscr{} incorporates symptoms on top of text ($\mathcal{P}(D_0\mid B,S_0,S_1,S_2,T)$), it improves over \ff{} with almost 10 percentage points in average precision, proving the merit of learning separate symptom classifiers, as well as diagnosis classifiers, and incorporating all their predictions through a Bayesian inference procedure. 


Table \ref{tab:res_inf} in Appendix \ref{sec:app:res} shows higher average precision measures for the prediction of $inf$, which is much more common in our dataset than $pneu$. 
Note that including the text in the prediction does not improve performance. Indeed, the small gap between \bn{} and \bnplus{} shows that knowledge on the symptoms $fever$ and $pain$ does not improve the prediction of $inf$ much. Their textual representation is therefore expected to have little impact as well.

\begin{table}[t]
    \caption{Average precision over test set of three posterior probabilities for the diagnosis $D_0$ ($pneu$), each taking a different set of evidence into account (various combinations of background, symptoms and text). 
    We show mean ($\pm$ std) over 5 initialization seeds.}
    \label{tab:res_pneu}
    \begin{tabularx}{\textwidth}{
    >{\hsize=0.72\hsize}X 
    >{\hsize=1.12\hsize\centering\arraybackslash}X 
    >{\hsize=1.08\hsize\centering\arraybackslash}X 
    >{\hsize=1.08\hsize\centering\arraybackslash}X 
    }
        \toprule
        & \multicolumn{3}{c}{\textbf{Average precision for \textit{pneu}}} \\
        \textbf{Model} & \textbf{$P(D_0\mid B,S_0,S_1,S_2,T)$} & \textbf{$P(D_0\mid B,S_0,S_1,S_2)$} & \textbf{$P(D_0\mid B,T)$} \\
        \hline
        \bn{} & - & 0.0914 ($\pm$ 0.0000) & - \\
        \bnplus{} & - & \textbf{0.8326} ($\pm$ 0.0000) & - \\
        \mbox{\ff{}} & 0.6574 ($\pm$ 0.0118) & 0.1090 ($\pm$ 0.0020) & 0.6220 ($\pm$ 0.0121) \\
        \mbox{\bngen{}} & 0.5870 ($\pm$ 0.0000) & 0.0892 ($\pm$ 0.0007) & 0.4434 ($\pm$ 0.0000) \\
        \mbox{\bndiscr{}} & \textbf{0.7538} ($\pm$ 0.0323) & 0.1079 ($\pm$ 0.0011) & \textbf{0.6922} ($\pm$ 0.0273) \\
        \bottomrule
    \end{tabularx}
\shrink \shrink \shrink
\end{table}

\shrink
\section{Discussion and Conclusion} \label{sec:conclusion} 
\shrink

We conclude with a discussion of how the results from the previous section can be seen in a broader context, by answering three key questions on the integration of text in Bayesian networks (BNs) for clinical reasoning.\\

\shrink
\noindent
\textbf{What are different ways to integrate text into a BN, to allow for joint reasoning over unstructured text and structured tabular features? 
}

We compared two architectures belonging to complementary model families: a BN with text generator (\bngen{}) and a BN with text discriminator (\bndiscr{}). An advantage of the \bngen{} model is that it follows the causal structure of the text generation process, making it more intuitive to understand. However, to fit a generative model for the text embeddings, we need to make assumptions on the distribution which do not hold in practice. 
Conditional fitting of the Gaussians for every combination of diagnoses and symptoms also leads to a bad fit for rarer combinations. Both of these downsides translate to inferior performance of the generative model on our use case. However, alternative generative architectures are worth exploring in future research.
The \bndiscr{} model can benefit from the flexibility of the neural classifiers without requiring
any assumptions on the distribution of the text embeddings. 
Its modular approach (with separate classifiers for the diagnoses and symptoms) allows for an intuitive integration of the observed symptoms into the Bayesian inference procedure. \\

\shrink
\noindent
\textbf{What are the advantages of using unstructured text for clinical reasoning, compared to only using structured tabular features?}

Reducing clinical text to a set of structured variables can be challenging, 
and inherently causes loss of information. By retaining the raw text and training the model to deal with this, the information extraction step is no longer required.  This avoids the need to determine up front which variables are (1) relevant for any set of diagnoses, (2) mentioned frequently enough, and (3) can be extracted with sufficient accuracy. We simulated the presence of complementary information in the text with the symptoms $fever$ and $pain$.
The models that included the text, were able
to leverage information on those symptoms without explicitly including them as variables in the BN, which turned out especially beneficial for the rarer disease (pneumonia). 
%
This aligns with the intuition that specific symptoms related to rare diseases may not be encoded due to their infrequent presence,
while at the same time being indispensable for accurate diagnosis. \\


\shrink
\noindent
\textbf{What are the advantages of Bayesian inference for clinical reasoning, compared to approaches that don’t contain a BN component?} 

BNs 
model each conditional distribution separately. This is not the case for the \ff{} baseline, which directly outputs a prediction for the diagnosis instead. This modular approach has multiple advantages. First of all, it helps the model deal with missing data during the training process: conditional distributions for variables that are not observed in a particular sample are simply not updated. 
The \ff{} baseline deals with unobserved symptoms by incorporating a special category in its one-hot encoding, which is much less natural. 
Second, and even more important, this modularity improves the interpretability of the prediction, which is essential in medical applications. An end user of the \bndiscr{} model can inspect the outputs of the symptom classifiers as well as the diagnosis classifier, and see how all these probabilities contribute to the predicted posterior through the Bayesian inference process. 

\shrink
\section*{Acknowledgements}
\shrink

Paloma Rabaey’s research is funded by the Research Foundation Flanders (FWO-Vlaanderen) with grant number 1170122N. This research also received funding from the Flemish government under the “Onderzoeksprogramma Artificiële Intelligentie (AI) Vlaanderen”
programme.

\bibliographystyle{splncs04}
\shrink\shrink
\bibliography{bibliography}

\begin{thebibliography}{10}
\providecommand{\url}[1]{\texttt{#1}}
\providecommand{\urlprefix}{URL }
\providecommand{\doi}[1]{https://doi.org/#1}

\bibitem{pgmpy}
{A}nkur {A}nkan, {A}binash {P}anda: pgmpy: {P}robabilistic {G}raphical {M}odels using {P}ython. In: {P}roceedings of the 14th {P}ython in {S}cience {C}onference. pp. 6 -- 11 (2015)

\bibitem{clinical_judgement}
Chin-Yee, B., Upshur, R.: {{C}linical judgement in the era of big data and predictive analytics}. J Eval Clin Pract  \textbf{24}(3),  638--645 (2018)

\bibitem{precision_recall_ROC}
Davis, J., Goadrich, M.: The relationship between precision-recall and roc curves. In: Proceedings of the 23rd International Conference on Machine Learning. p. 233–240 (2006)

\bibitem{BN_respiratory}
Edye, E.O., Kurucz, J.F., Lois, L., Paredes, A., Piria, F., Rodríguez, J., Delgado, S.H.: Applying {B}ayesian networks to help physicians diagnose respiratory diseases in the context of covid-19 pandemic. In: 2021 IEEE URUCON. pp. 368--371 (2021)

\bibitem{extracting_information_EMR}
Ford, E., Carroll, J.A., Smith, H.E., Scott, D., Cassell, J.A.: {Extracting information from the text of electronic medical records to improve case detection: a systematic review}. J Am Med Inform Assoc  \textbf{23}(5),  1007--1015 (2016)

\bibitem{clinical_reasoning}
Gruppen, L.D.: {C}linical reasoning: Defining it, teaching it, assessing it, studying it. West J Emerg Med  \textbf{18}(1), ~4--7 (2017)

\bibitem{BNadoption}
Kyrimi, E., Dube, K., Fenton, N., Fahmi, A., Neves, M.R., Marsh, W., McLachlan, S.: Bayesian networks in healthcare: What is preventing their adoption? Artif Intell Med  \textbf{116},  102079 (2021)

\bibitem{kyrimi_scoping_BN}
Kyrimi, E., McLachlan, S., Dube, K., Neves, M.R., Fahmi, A., Fenton, N.: A comprehensive scoping review of {B}ayesian networks in healthcare: Past, present and future. Artif Intell Med  \textbf{117},  102108 (2021)

\bibitem{Manhaeve2019_DPL}
Manhaeve, R., Dumancic, S., Kimmig, A., Demeester, T., De~Raedt, L.: Deep{P}rob{L}og: Neural probabilistic logic programming. In: Advances in Neural Information Processing Systems (NeurIPS). vol.~31 (2018)

\bibitem{Marra2024_NeSyAI}
Marra, G., Duman{\v c}i{\'c}, S., Manhaeve, R., {De Raedt}, L.: From statistical relational to neurosymbolic artificial intelligence: A survey. Artif Intell  \textbf{328},  104062 (2024)

\bibitem{BN_healthcare_condition}
McLachlan, S., Dube, K., Hitman, G.A., Fenton, N.E., Kyrimi, E.: Bayesian networks in healthcare: Distribution by medical condition. Artif Intell Med  \textbf{107},  101912 (2020)

\bibitem{clinical_text_classification}
Mujtaba, G., Shuib, L., Idris, N., Hoo, W.L., Raj, R.G., Khowaja, K., Shaikh, K., Nweke, H.F.: Clinical text classification research trends: Systematic literature review and open issues. Expert Syst Appl  \textbf{116},  494--520 (2019)

\bibitem{clinical_reasoning_errors}
Norman, G.R., Monteiro, S.D., Sherbino, J., Ilgen, J.S., Schmidt, H.G., Mamede, S.: {T}he causes of errors in clinical reasoning: Cognitive biases, knowledge deficits, and dual process thinking. Acad Med  \textbf{92}(1),  23--30 (2017)

\bibitem{instructGPT}
Ouyang, L., Wu, J., Jiang, X., Almeida, D., et~al.: Training language models to follow instructions with human feedback. In: Advances in Neural Information Processing Systems. vol.~35, pp. 27730--27744 (2022)

\bibitem{CDS_infectious_diseases}
Peiffer-Smadja, N., Rawson, T., Ahmad, R., Buchard, A., et~al.: Machine learning for clinical decision support in infectious diseases: {A} narrative review of current applications. Clin Microbiol Infect  \textbf{26}(5),  584--595 (2020)

\bibitem{omission_free_text}
Price, S.J., Stapley, S.A., Shephard, E., Barraclough, K., Hamilton, W.T.: Is omission of free text records a possible source of data loss and bias in clinical practice research datalink studies? {A} case{\textendash}control study. BMJ Open  \textbf{6}(5),  e011664 (2016)

\bibitem{BioLORD}
Remy, F., Demuynck, K., Demeester, T.: {B}io{LORD}: Semantic textual representations fusing {LLM} and clinical knowledge graph insights. arXiv preprint  (2023)

\bibitem{health_knowledge_graph}
Rotmensch, M., Halpern, Y., Tlimat, A., Horng, S., Sontag, D.: Learning a health knowledge graph from electronic medical records. Scientific reports  \textbf{7}(1), ~5994 (2017)

\bibitem{sterckx2020_clinical_IE}
Sterckx, L., Vandewiele, G., Dehaene, I., Janssens, O., et~al.: Clinical information extraction for preterm birth risk prediction. J Biomed Inform  \textbf{110},  103544 (2020)

\bibitem{evidence_based_medicine}
Strauss, S.E., Glasziou, P., Richardson, W.S., Haynes, R.B.: Evidence-based medicine. {H}ow to practice and teach {EBM}., vol.~5. Elsevier (2018)

\bibitem{computer_aided_diagnosis}
Yanase, J., Triantaphyllou, E.: A systematic survey of computer-aided diagnosis in medicine: Past and present developments. Expert Syst Appl  \textbf{138},  112821 (2019)

\bibitem{clinical_reasoning_yazdani}
Yazdani, S., Hoseini~Abardeh, M.: {{F}ive decades of research and theorization on clinical reasoning: {A} critical review}. Adv Med Educ Pract  \textbf{10},  703--716 (2019)

\bibitem{clinical_reasoning_GP}
Yazdani, S., Hosseinzadeh, M., Hosseini, F.: {{M}odels of clinical reasoning with a focus on general practice: {A} critical review}. J Adv Med Educ Prof  \textbf{5}(4),  177--184 (2017)

\bibitem{influenza_detection}
Ye, Y., Tsui, F., Wagner, M., Espino, J., Li, Q.: {Influenza detection from emergency department reports using natural language processing and {B}ayesian network classifiers}. J Am Med Inform Assoc  \textbf{21}(5),  815--823 (2014)

\bibitem{combining_structured_unstructured}
Zhang, D., Yin, C., Zeng, J., Yuan, X., Zhang, P.: {{C}ombining structured and unstructured data for predictive models: {A} deep learning approach}. BMC Med Inform Decis Mak  \textbf{20}(1), ~280 (2020)

\end{thebibliography}
\shrink\shrink\shrink






\appendix
\clearpage
\newpage

\setcounter{table}{0}
\renewcommand{\thetable}{A\arabic{table}}
\setcounter{figure}{0}
\renewcommand{\thefigure}{A\arabic{figure}}

\section*{Appendix}


\shrink
\section{Data Generation Process} 
\label{sec:app:data_gen}
\shrink

Figure \ref{fig:data_generation} in the main text shows the steps we take to generate our dataset. 

\begin{figure}[t]
    \centering
    \includegraphics[width=0.8\textwidth]{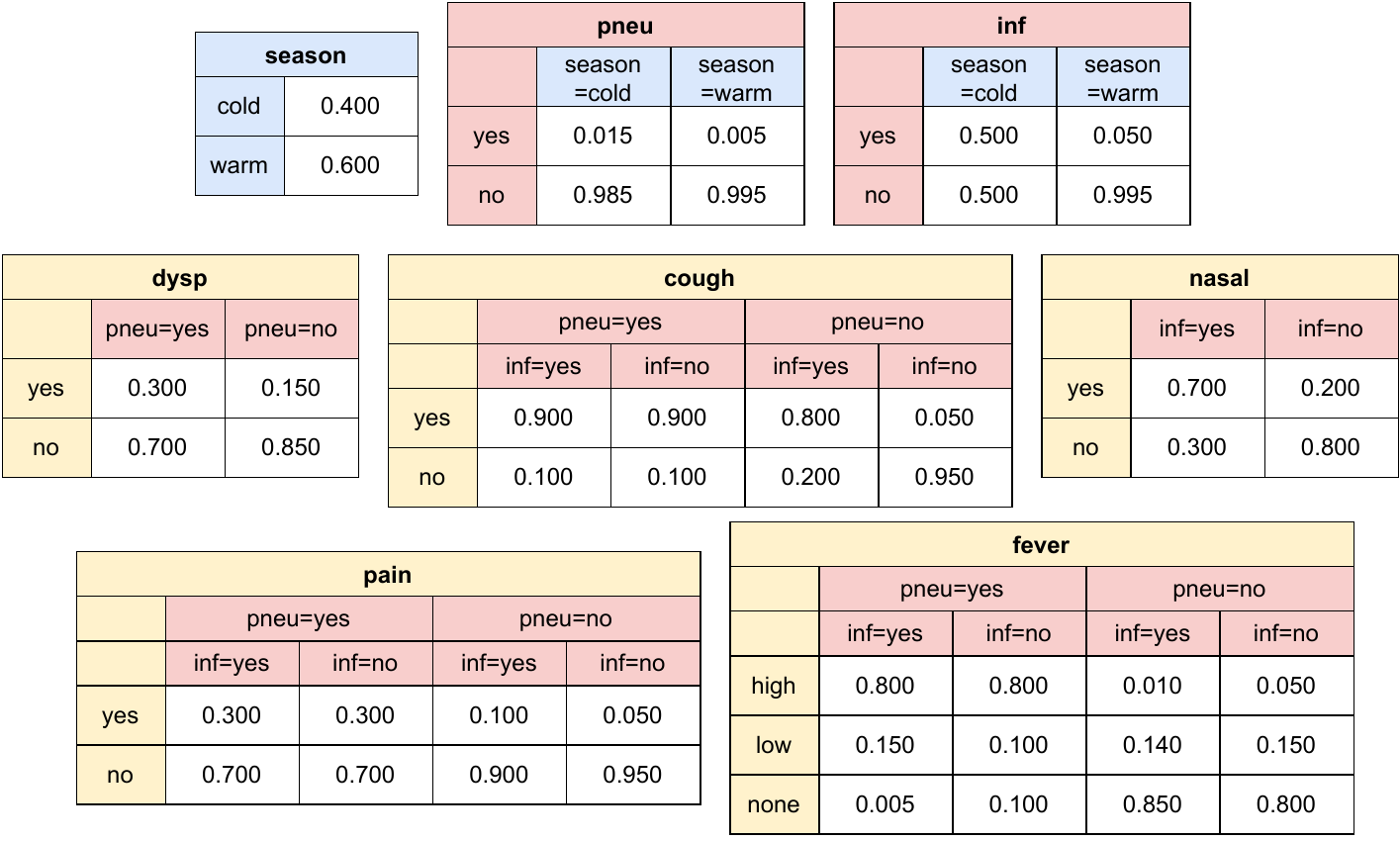}
    \caption{Conditional probability tables (CPTs) for all parent-child relations in the ground truth Bayesian network, as defined by an expert general practitioner. 
    }
    \label{fig:cpts}
\end{figure}

\textbf{Step 1. Expert-defined Bayesian network}\quad With help of an expert general practitioner, we define a Bayesian network (BN) that can be used for diagnosis of two diseases: pneumonia ($pneu$) and upper respiratory tract infection ($inf$). 
We model the effect of one background factor, $season$ of the year, on both diagnoses. Additionally, five symptoms are added as nodes to the network: $dysp$ (dyspnea, also known as shortness of breath), $cough$, $fever$, $pain$ (chest pain and/or pain attributed to airways) and $nasal$ (nasal congestion and/or sneezing). All variables are binary ($warm$/$cold$ for $season$ and $no$/$yes$ for all others), except for $fever$, which can take on three levels ($none$/$low$/$high$). According to the expert, these five symptoms are the most informative to diagnose the two respiratory diseases in a primary care setting. Figure \ref{fig:cpts} shows the conditional probability tables (CPTs, defining the distribution of any child conditioned on its possible parent values), which were filled in according to the expert's own knowledge and experience. The product of all conditional distributions forms the joint ground truth distribution $\mathcal{P}_{GT}$ modeled by the BN, as shown in Equation \ref{eq:joint_GT}.

\shrink\shrink\shrink
\begin{multline} \label{eq:joint_GT}
\mathcal{P}_{GT}(season,pneu,inf,dysp,cough,fever,pain,nasal)
= \mathcal{P}_{GT}(season)\\
\mathcal{P}_{GT}(pneu\mid season)\mathcal{P}_{GT}(inf\mid season)\mathcal{P}_{GT}(dysp\mid pneu)
\mathcal{P}_{GT}(cough\mid pneu,inf)\\
\mathcal{P}_{GT}(fever\mid pneu,inf)\mathcal{P}_{GT}(pain\mid pneu,inf)\mathcal{P}_{GT}(nasal\mid inf)
\end{multline}

\textbf{Step 2. Sample tabular variables}\quad We can easily sample from the joint distribution $\mathcal{P}_{GT}$ in a top-down fashion, starting from the root node $season$, generating the 2 diagnoses conditioned on $season$, and finally generating the 5 symptoms conditioned on the diagnoses. This way, we obtain 4000 train samples and 1000 test samples, each with 8 tabular features. We use the \textit{pgmpy} library for implementing our BN and sampling from it \cite{pgmpy}.

\textbf{Step 3. Generate clinical consultation notes}\quad We then prompt a large language model (LLM, in our case the GPT-3.5-turbo model from OpenAI \cite{instructGPT}) to generate textual descriptions for each sample, given the presence or absence of the tabular symptoms. We want these textual descriptions to resemble clinical consultation notes made by a general practitioner for each fictitious patient encounter, which means the LLM only gets to observe the symptoms, not the diagnoses. 
Appendix \ref{sec:app:data_gen_prompt} outlines the full prompting strategy.

\setlength{\belowcaptionskip}{-10pt}
\setlength{\abovecaptionskip}{2pt}

\textbf{Step 4. Make data partially observed}\quad While our dataset now fulfills desired properties (i) and (iii) as outlined in Section \ref{sec:data_gen} in the main text, we still need to enforce property (ii), which we split up into 3 parts. 
\shrink
\begin{itemize}
    \item \textbf{Step 4a:} Some symptoms are never encoded in the tabular portion of the data at all. To mimic this, we completely remove features $fever$ and $pain$ from the dataset. This way, none of our models (except for baseline model \bnplus{}, see later) ever observe these variables directly in tabular format, instead only seeing indirect mentions of them in the text. Both the symptoms $fever$ and $pain$ are highly informative for the prediction of pneumonia, and any model that can extract information from the text should reap the benefits.
    \item \textbf{Step 4b:} Other symptoms are only encoded in a subset of the data. We simulate this situation by masking out the remaining symptoms in a subset of the training data. For 1/3 of the training data (1333 samples), we leave out the values for variables $dysp$, $cough$ and $nasal$, rendering them unobserved. Note that we either observe all 3 symptoms or none at all, thereby avoiding the need to model missingness and simplifying reality. 
    \item \textbf{Step 4c:} Furthermore, real data might not contain textual descriptions for all samples. 
    For this reason, we mask out the textual description for another 1/3 of the training data (1333 samples). This leaves the remaining 1/3 of the training data (1334 samples), with fully observed symptoms and text. 
\end{itemize}
    
We assume that the background variable is always observed -- in a real setting, it could be extracted from the timestamp of the electronic record -- and therefore never mask it out. The diagnoses variables are never masked out either, mostly to simplify the setup. We don't mask out anything in the test set, to leave full flexibility during the evaluation process in deciding what to include as input to the predictive models. 
 

\shrink\shrink
\section{Prompting Strategies} \label{sec:app:data_gen_prompt}
\shrink

There are 5 symptoms in our tabular dataset, forming a total of 48 possible combinations. To mimic a realistic setting, we want each sample to have a unique textual description, meaning we need to generate a wide variety of different texts for each of these combinations of symptoms. We asked an expert general practitioner to provide us with some example patient encounter notes that could be used to prompt the LLM and encourage some variety in its responses. The expert received a symptom configuration and was asked to describe the patient encounter like they normally would in practice. We manually translated these notes from Dutch to English. We ensured that the top 10 most occurring symptom combinations in the training set (for which we will need to generate the highest number of unique textual descriptions) have at least one clinical example note. We had 20 annotated example notes in total. 
Since some of these notes are based on real encounters the general practitioner remembered from their own clinical practice, we do not make these public.

We now describe our LLM prompting strategy. We structure all prompts according to the OpenAI Chat Completions interface
with the GPT-3.5-turbo model, using a temperature of 1 and a frequency penalty of 0.5. The full code to reproduce our prompting strategy is available in our Github repository: \url{https://github.com/prabaey/bn-text}. 

Suppose we need to generate a text description for a symptom combination \{$dysp=d$, $cough=c$, $fever=f$, $pain=p$, $nasal=n$\} that needs $m$ unique textual descriptions. First, we check whether this combination is present in the set of examples. If one or more examples are found, we start our prompt with the requested symptom combination, followed by the example responses, see \url{https://platform.openai.com/playground/p/poSdvoy9dipYIwVPXepRrUyL?model=gpt-3.5-turbo&mode=chat}. 
If no example is found, we randomly pick two unrelated examples and prompt the language model by listing one after the other, preceded by their corresponding symptom combinations, see \url{https://platform.openai.com/playground/p/6KOm6pP6DXmMwDxUJWdGGHP1?model=gpt-3.5-turbo&mode=chat}. 
In both scenarios, after showing the examples, we ask the LLM to generate 5 clinical notes. We repeat the prompt as many times as needed to build up a set of $m$ notes. To further encourage diversity in the responses, we only mention symptoms with positive values in the prompt in 50\% of the cases, while in the other 50\% we mention all symptoms and their values. 
We execute the entire pipeline separately for the train and test set. A random sample of the resulting notes were checked for coherence and correctness by the authors, which were deemed sufficient for this proof-of-concept setting.

We use a separate prompting strategy for the combination where all symptoms are absent. This combination occurs most often out of all
, though we still only have 4 example notes for it. If we were to exclusively use the prompting strategy from scenario 1, the notes would have little variety. For this reason, we use 5 different strategies that each account for a different number of generated notes. While some strategies encourage the model to mention the absence of respiratory symptoms, others encourage the model to invent a completely unrelated patient encounter. We once again conduct the entire process for the train and test set separately, with the train set needing $1032$ descriptions and the test set needing $388$. We describe the 5 strategies we used, together with the proportion of textual descriptions we generated using each strategy and how many samples this comes down to in both the train and test set. These proportions were decided arbitrarily based on how useful we deemed each strategy to be.
\shrink
\begin{enumerate}
    \item Provide two in-context examples for symptom combination \{$dysp=no$, $cough=no$, $fever=none$, $pain=no$, $nasal=no$\}: \url{https://platform.openai.com/playground/p/Zw9y4EZ8RRfGaZTgBCPu5DPT?model=gpt-3.5-turbo&mode=chat}. ($40\%$, Train: $\#400$, Test: $\#151$)
    \item Provide two random out-of-context examples for other symptom combinations: \url{https://platform.openai.com/playground/p/jYJeFfzXVZ5akndHGkB6cmPk?model=gpt-3.5-turbo&mode=chat}. ($5\%$, Train: $\#50$, $\#19$)
    \item Similar to strategy 1, but do not mention the absent symptoms explicitly, thereby encouraging the model to describe cases outside of the respiratory domain: \url{https://platform.openai.com/playground/p/Ikszv18Zbqr162kF0iSErbCT?model=gpt-3.5-turbo&mode=chat}. We manually go over the generated cases and filter those out where $dysp$, $cough$, $fever$, $pain$ and $nasal$ are described as present in the patient. ($10\%$, Train: $\#107$, Test: $\#39$)
    \item Same as strategy 3, but don't show any examples: \url{https://platform.openai.com/playground/p/8KaVpr7chxLMHyeKz6Y2mMVE?model=gpt-3.5-turbo&mode=chat}. We manually go over the generated cases and filter those out where $dysp$, $cough$, $fever$, $pain$ and $nasal$ are described as present in the patient. ($10\%$, Train: $\#111$, Test: $\#43$)
    \item Same as stragegy 4, still without showing any examples, but this time telling the model that the patient is not experiencing the symptoms $dysp$, $cough$, $fever$, $pain$ and $nasal$: \url{https://platform.openai.com/playground/p/eXDqpwMkqcvUA8wWiW1H06et?model=gpt-3.5-turbo&mode=chat}. 
    ($35\%$, Train: $\#364$, Test: $\#136$) \shrink\shrink
\end{enumerate}

\shrink
\section{Augmenting BNs with Text Representations} \label{sec:app:models}
\shrink

\subsection{Baseline: Bayesian Network (\bn{} and \bnplus{})} \label{sec:model_BN}
\shrink

\textbf{Training} \quad We train a simple Bayesian network (\bn{}) where the Directed Acyclic Graph (DAG), which defines the structure between all the tabular variables, is the same as the one used to generate the data (see Figure \ref{fig:data_generation}, excluding the unobserved symptoms). This Bayesian network defines the joint distribution as a product of six conditional distributions, one for each variable, as shown in equation \ref{eq:joint_BN}. These distributions are learned from the training data using maximum likelihood estimation. This method studies the co-occurrence of particular values of each variable and its parents in the training set, filling up the CPTs as such. We use a K2 prior as a smoothing strategy, to counteract the extremely skewed probability distributions that might be learned when particular combinations of variables are never observed in the training set. We use the \textit{pgmpy} Python library to learn the Bayesian network \cite{pgmpy}. 

\shrink\shrink\shrink
\begin{multline} \label{eq:joint_BN}
\mathcal{P}(B,D_0,D_1,S_0,S_1,S_2) = \mathcal{P}(B)\mathcal{P}(D_0\mid B)\mathcal{P}(D_1\mid B)\mathcal{P}(S_0\mid D_0)\\
\mathcal{P}(S_1\mid D_0,D_1)\mathcal{P}(S_2\mid D_1)
\end{multline}
\shrink

\shrink\shrink
\noindent
\textbf{Inference} \quad The baseline Bayesian network can only include background and symptoms as evidence (no text). We can calculate the posterior for either diagnosis $D_i$ by applying Bayes' rule and performing marginalization over the variables which are not included in the evidence, as shown in Equation \ref{eq:inf_BN}. We use the Variable Elimination method from \textit{pgmpy} to perform exact inference. 

\shrink
\begin{equation} 
\label{eq:inf_BN}
\mathcal{P}(D_i\mid B,S_0,S_1,S_2) = \frac{\sum_{D_{1-i}} \mathcal{P}(B,D_0,D_1,S_0,S_1,S_2)}{\sum_{D_0,D_1}\mathcal{P}(B,D_0,D_1,S_0,S_1,S_2)}
\end{equation}
\shrink

\shrink
\noindent
\textbf{Inclusion of unobserved symptoms} \quad We also build a second variant of this baseline (\bnplus{}) where we additionally include the unobserved symptoms $fever$ and $pain$ in the DAG. As opposed to all other models, this model is trained on a version of the training data where these two variables are \textit{not} masked out. This baseline serves as an upper bound to the performance of all other models, which never get to directly observe these two symptoms. Equation \ref{eq:joint_BNplus} shows modeled the joint distribution, where $U_0$ and $U_1$ represent the unobserved symptoms $fever$ and $pain$ respectively. Equation \ref{eq:inf_BN} can be trivially extended to obtain $\mathcal{P}(D_i\mid B,S_0,S_1,S_2,U_0,U_1)$, where evidence does not only include the partially observed symptoms $S_i$, but also the unobserved symptoms $U_i$.

\shrink\shrink\shrink
\begin{multline} \label{eq:joint_BNplus}
\mathcal{P}(B,D_0,D_1,S_0,S_1,S_2,U_0,U_1) = \mathcal{P}(B)\mathcal{P}(D_0\mid B)\mathcal{P}(D_1\mid B)\mathcal{P}(S_0\mid D_0) \\
\mathcal{P}(S_1\mid D_0,D_1)\mathcal{P}(S_2\mid D_1)\mathcal{P}(U_0\mid D_0,D_1)\mathcal{P}(U_1\mid D_0,D_1)
\end{multline}
\shrink\shrink\shrink

\shrink
\subsection{Baseline: Feed-Forward Neural Network (\ff{})} \label{sec:model_FF}
\shrink

\textbf{Training} \quad We train two discriminative feed-forward neural networks (\ff{}) which receive a vector representation of both the tabular features and the text at the input, and transform it into a one-dimensional representation which is turned into a prediction for $\mathcal{P}(D_i\mid B,S_0,S_1,S_2,T)$ by applying a $sigmoid$ non-linearity. We build two completely separate models, one for $pneu$ and one for $inf$, and optimize the neural network weights using maximum-likelihood estimation with a binary cross-entropy loss. As a vector representation for the text, we use \textit{BioLORD} \cite{BioLORD}, which returns a 768-dimensional embedding of the text description. The tabular features are turned into a one-hot encoding, with 11 dimensions in total. Note that each symptom is encoded into a three-dimensional vector, to be able to model the case where the symptom is unobserved, next to its two possible classes ($yes$/$no$). We also experimented with including pairwise, three-way and four-way interactions of background and symptom representations at the input, which adds another 180 dimensions. 
See Appendix \ref{sec:app:training} for the final hyperparameter configuration.

\noindent
\textbf{Inference} \quad The model is trained to maximize the likelihood $\mathcal{P}(D_i\mid B,\allowbreak S_0, \allowbreak S_1, \allowbreak S_2,\allowbreak T)$, so we can directly obtain this probability as an output to the model when we input a test sample. To get a prediction for $\mathcal{P}(D_i\mid B,S_0,S_1,S_2)$, we replace the text at the input by an empty string (simply ``") and use its \textit{BioLORD} embedding. Note that the model is equipped to deal with this, since these empty texts occur in 1/3 of the training data as well. Finally, to get a prediction for $\mathcal{P}(D_i\mid B,T)$, we set all symptoms to unobserved and use their corresponding one-hot encoding at the input of the model, instead of the original encoding. 

\shrink
\subsection{Bayesian Network with Text Generator (\bngen{})} \label{sec:app:gen}
\shrink

The joint probability distribution modeled by the Bayesian network with text generator is given in Equation \ref{eq:joint_gen}. We parameterize each conditional distribution as a Bernoulli distribution with one trainable parameter per conditional parent configuration, except for the text $T$, which fits a Gaussian distribution to the text embeddings as explained in Section \ref{sec:model_gen}. We then learn all trainable parameters using maximum likelihood estimation where the likelihood $\mathcal{P}(B,D_0,D_1,S_0,S_1,S_2)$ (Equation \ref{eq:joint_gen} without factor $\mathcal{P}(T\mid D_0,D_1,S_0,S_1,S_2)$) is maximized based on the training data. This essentially comes down to filling in the CPTs
like in a normal Bayesian network. 

\shrink\shrink\shrink
\begin{multline} \label{eq:joint_gen}
\mathcal{P}(B,D_0,D_1,S_0,S_1,S_2,T) = \mathcal{P}(B)\mathcal{P}(D_0\mid B)\mathcal{P}(D_1\mid B)\mathcal{P}(S_0\mid D_0)\\
\mathcal{P}(S_1\mid D_0,D_1)\mathcal{P}(S_2\mid D_1)\mathcal{P}(T\mid D_0,D_1,S_0,S_1,S_2)
\end{multline}

Note that the generative model can easily deal with missing data: if the symptoms are unobserved, only the parameters for $\mathcal{P}(B)$, $\mathcal{P}(D_0\mid B)$ and $\mathcal{P}(D_1\mid B)$ are updated. Similarly, samples where the text is missing still contribute to the learned CPTs, while $\mathcal{P}(T\mid D_0,D_1,S_0,S_1,S_2)$ is fitted separately to the observed text embeddings only. 

Equations \ref{eq:gen_inf_sympt}, \ref{eq:gen_inf_sympt_text} and \ref{eq:gen_inf_text} show how we calculate the posterior likelihood for the diagnoses through Bayesian inference, for different sets of evidence. Note that we write sums for clarity, but strictly speaking marginalization over $T$ is done by integration over the normally distributed text embedding variable.

\shrink\shrink\shrink
\begin{multline} \label{eq:gen_inf_sympt}
\mathcal{P}(D_i\mid B,S_0,S_1,S_2) = \frac{\sum_{D_{1-i},T} \mathcal{P}(B,D_0,D_1,S_0,S_1,S_2,T)}{\sum_{D_0,D_1,T}\mathcal{P}(B,D_0,D_1,S_0,S_1,S_2,T)} \\
= \frac{\splitfrac{\sum_{D_{1-i}} \mathcal{P}(B)\mathcal{P}(D_0\mid B)\mathcal{P}(D_1\mid B)\mathcal{P}(S_0\mid D_0)\mathcal{P}(S_1\mid D_0,D_1)}{\mathcal{P}(S_2\mid D_1)\sum_{T}\mathcal{P}(T\mid D_0,D_1,S_0,S_1,S_2)}}{\splitfrac{\sum_{D_0,D_1}\mathcal{P}(B)\mathcal{P}(D_0\mid B)\mathcal{P}(D_1\mid B)\mathcal{P}(S_0\mid D_0)\mathcal{P}(S_1\mid D_0,D_1)}{\mathcal{P}(S_2\mid D_1)\sum_{T}\mathcal{P}(T\mid D_0,D_1,S_0,S_1,S_2)}} \\
= \frac{\sum_{D_{1-i}} \mathcal{P}(B)\mathcal{P}(D_0\mid B)\mathcal{P}(D_1\mid B)\mathcal{P}(S_0\mid D_0)\mathcal{P}(S_1\mid D_0,D_1)\mathcal{P}(S_2\mid D_1)}{\sum_{D_0,D_1}\mathcal{P}(B)\mathcal{P}(D_0\mid B)\mathcal{P}(D_1\mid B)\mathcal{P}(S_0\mid D_0)\mathcal{P}(S_1\mid D_0,D_1)\mathcal{P}(S_2\mid D_1)}
\end{multline}\shrink

\shrink\shrink\shrink
\begin{multline} \label{eq:gen_inf_sympt_text}
\mathcal{P}(D_i\mid B,S_0,S_1,S_2,T) = \frac{\sum_{D_{1-i}} \mathcal{P}(B,D_0,D_1,S_0,S_1,S_2,T)}{\sum_{D_0,D_1}\mathcal{P}(B,D_0,D_1,S_0,S_1,S_2,T)} \\
= \frac{\splitfrac{\sum_{D_{1-i}} \mathcal{P}(B)\mathcal{P}(D_0\mid B)\mathcal{P}(D_1\mid B)\mathcal{P}(S_0\mid D_0)\mathcal{P}(S_1\mid D_0,D_1)}{\mathcal{P}(S_2\mid D_1)\mathcal{P}(T\mid D_0,D_1,S_0,S_1,S_2)}}{\splitfrac{\sum_{D_0,D_1}\mathcal{P}(B)\mathcal{P}(D_0\mid B)\mathcal{P}(D_1\mid B)\mathcal{P}(S_0\mid D_0)\mathcal{P}(S_1\mid D_0,D_1)}{\mathcal{P}(S_2\mid D_1)\mathcal{P}(T\mid D_0,D_1,S_0,S_1,S_2)}}
\end{multline}

\shrink\shrink\shrink
\begin{multline} \label{eq:gen_inf_text}
\mathcal{P}(D_i\mid B,T) = \frac{\sum_{D_{1-i},S_0,S_1,S_2} \mathcal{P}(B,D_0,D_1,S_0,S_1,S_2,T)}{\sum_{D_0,D_1,S_0,S_1,S_2}\mathcal{P}(B,D_0,D_1,S_0,S_1,S_2,T)} \\
= \frac{\splitfrac{\sum_{D_{1-i},S_0,S_1,S_2} \mathcal{P}(B)\mathcal{P}(D_0\mid B)\mathcal{P}(D_1\mid B)\mathcal{P}(S_0\mid D_0)\mathcal{P}(S_1\mid D_0,D_1)}{\mathcal{P}(S_2\mid D_1)\mathcal{P}(T\mid D_0,D_1,S_0,S_1,S_2)}}{\splitfrac{\sum_{D_0,D_1,S_0,S_1,S_2}\mathcal{P}(B)\mathcal{P}(D_0\mid B)\mathcal{P}(D_1\mid B)\mathcal{P}(S_0\mid D_0)\mathcal{P}(S_1\mid D_0,D_1)}{\mathcal{P}(S_2\mid D_1)\mathcal{P}(T\mid D_0,D_1,S_0,S_1,S_2)}}
\end{multline}
\shrink

\shrink
\subsection{Bayesian Network with Text Discriminator (\bndiscr{})} \label{sec:app:discr}
\shrink

The joint probability distribution modeled by the Bayesian network with text discriminator is given by Equation \ref{eq:joint_discr}. Like before, $\mathcal{P}(B)$ is parameterized as a Bernoulli distribution with one trainable parameter. All other conditional probability distributions are parameterized by one neural text discriminator per conditional parent configuration, resulting in 2 classifiers per conditional distribution, except for $\mathcal{P}(S_1\mid D_0,D_1,T)$, which has 4 (due to $cough$ having both $pneu$ and $inf$ as a parent). Each text classifier is modeled as a discriminative feed-forward neural network that takes a vector representation of the text as an input (once again, we use 768-dimensional \textit{BioLORD} embeddings), transforming it into a one-dimensional representation which is then turned into a prediction for $\mathcal{P}(X\mid \mathcal{Y}=y,T)$ by applying a \textit{sigmoid} non-linearity. Here, $X$ represents the child variable (either $D_0$, $D_1$, $S_0$, $S_1$ or $S_2$), while $\mathcal{Y}$ represents the parent variable (either $B$, $D_0$, $D_1$ or $\{D_0,D_1\}$) taking on the configuration $y$. The Bernoulli parameter and neural network weights are trained by jointly maximizing the likelihood in Equation \ref{eq:joint_discr} over the training set. 

\shrink\shrink\shrink
\begin{multline} \label{eq:joint_discr}
\mathcal{P}(B,D_0,D_1,S_0,S_1,S_2\mid T)
= \mathcal{P}(B)\mathcal{P}(D_0\mid B,T)\mathcal{P}(D_1\mid B,T)\mathcal{P}(S_0\mid D_0,T) \\
\mathcal{P}(S_1\mid D_0,D_1,T)\mathcal{P}(S_2\mid D_1,T)
\end{multline}
\shrink\shrink

Note that the discriminative model can easily deal with missing data: if the symptoms are unobserved, only the parameters for $\mathcal{P}(B)$, $\mathcal{P}(D_0\mid B,T)$ and $\mathcal{P}(D_1\mid B,T)$ are updated. When the text is unobserved, we input the \textit{BioLORD} embedding for an empty text into each classifier. For that particular input, the classifiers will simply learn the co-occurrence of child and parent values in the 1/3 of the training data where text is empty, their outputs essentially mimicking the CPTs in a normal Bayesian network (like in the \bn{} baseline model).

Equations \ref{eq:inf_discr} and \ref{eq:inf_discr_sympt} show how we calculate the posterior likelihood for the diagnoses through Bayesian inference, taking the symptoms and text (empty or not) as evidence. $\mathcal{P}(D_i\mid B,T)$ can simply be taken directly as the output of the relevant diagnosis classifier. 

\shrink\shrink\shrink
\begin{multline} \label{eq:inf_discr}
\mathcal{P}(D_i\mid B,S_0,S_1,S_2,T) = \frac{\sum_{D_{1-i}} \mathcal{P}(B,D_0,D_1,S_0,S_1,S_2\mid T)}{\sum_{D_0,D_1}\mathcal{P}(B,D_0,D_1,S_0,S_1,S_2\mid T)} \\
= \frac{\splitfrac{\sum_{D_{1-i}} \mathcal{P}(B)\mathcal{P}(D_0\mid B,T)\mathcal{P}(D_1\mid B,T)\mathcal{P}(S_0\mid D_0,T)}{\mathcal{P}(S_1\mid D_0,D_1,T)\mathcal{P}(S_2\mid D_1,T)}}{\splitfrac{\sum_{D_0,D_1}\mathcal{P}(B)\mathcal{P}(D_0\mid B,T)\mathcal{P}(D_1\mid B,T)\mathcal{P}(S_0\mid D_0,T)}{\mathcal{P}(S_1\mid D_0,D_1,T)\mathcal{P}(S_2\mid D_1,T)}}
\end{multline}
\shrink

\shrink
\begin{equation} \label{eq:inf_discr_sympt}
\mathcal{P}(D_i\mid B,S_0,S_1,S_2) = \mathcal{P}(D_i\mid B,S_0,S_1,S_2,T=``")
\end{equation}
\shrink

\shrink\shrink\shrink
\section{Empirical Results and Analysis} \label{sec:app:res}
\shrink

\subsection{Training, Hyperparameter Tuning and Evaluation} \label{sec:app:training}
\shrink

All models, except the baselines \bn{} and \bnplus{}, have multiple hyperparameters to tune. We optimized these separately for each model using a train-validation split on the train set (3200/800 sample split out of 4000 samples in total), choosing the hyperparameters that maximize the average precision of $\mathcal{P}(D_i|B,S_0,S_1,S_2,T)$ on the validation set. The full implementation can be found in our Github repository: \url{https://github.com/prabaey/bn-text}. \\

\shrink
\begin{itemize}
    \item \ff{} \quad We optimized the number of epochs, the number of layers (including their width), the batch size, learning rate and weight decay of the Adam optimizer, 
    dropout and whether to include interaction features at the input or not. We optimized these hyperparameters separately for the $pneu$ and $inf$ classifier. For the $pneu$ classifier, the final configuration we landed on was the following: $200$ epochs, 2 layers (dimensions $768 \rightarrow 256 \rightarrow 1$, with a ReLU activation in the middle), batch size $256$, learning rate $1\mathrm{e}{-2}$, weight decay $1\mathrm{e}{-3}$, dropout of $70\%$ in every layer and no interaction features. For the $inf$ classifier, the optimal settings were the same, except that it had 1 layer (dimensions $768 \rightarrow 1$). To make up for the lower complexity of the model (and limited ability to mix features in a single layer), it proved optimal to include the interaction features at the input of this classifier. 
    \item \bndiscr{} \quad To make for a fair comparison, we used the same layer and dropout configurations that were chosen after tuning the \ff{} model for the $\mathcal{P}(D_0|B,T)$ classifier ($pneu$) and $\mathcal{P}(D_1|B,T)$ classifier ($inf$). The symptom classifiers already achieved perfect performance with only 1 layer (dimensions $768 \rightarrow 1$) and without dropout, so we kept these settings. We again used the Adam optimizer to learn the neural weights for all classifiers, with learning rate $1\mathrm{e}{-2}$ and weight decay $1\mathrm{e}{-3}$. We used a separate learning rate of $0.05$ (without weight decay) for learning the $\mathcal{P}(B)$ distribution, which is modeled with a single Bernoulli parameter. Other hyperparameters were also chosen in accordance with the \ff{} model: $200$ epochs and batch size $256$. 
    \item \bngen{} \quad For learning the conditional probability table parameters in the Bayesian network, we used an Adam optimizer with a learning rate of $0.05$ and no weight decay. We trained for $15$ epochs with a batch size of $256$. Hyperparameter $\alpha$, which regularizes the covariance matrix in Equation \ref{eq:gauss_gen}, was found to be optimal at $0.85$. We use the same $\alpha$ for all 32 Gaussians.
\end{itemize}
 
We trained all models with their optimal hyperparameter configurations over the train set of 4000 samples. We repeated this process 5 times, each time with a different initialization seed (except for the \bn{} baseline, which is deterministic). For each trained model, we calculated the three posterior diagnosis probabilities for all 1000 samples in the test set. We then obtained the average precision (area under the precision-recall curve) by comparing each prediction to the known label for the diagnosis. We report average precision rather than area under the ROC curve (another metric often used to assess classification performance), since the former is better suited to evaluate predictive performance in extremely imbalanced datasets \cite{precision_recall_ROC}, which is the case for pneumonia.\footnote{We have a positive pneumonia label for only 34 out of 4000 samples in the training set and 14 out of 1000 samples in the test set.} Furthermore, balancing precision and recall (catching as many cases of pneumonia as possible without including too many false positives) describes the diagnostic task of the GP in the practical use case well. Table \ref{tab:res_pneu} in the main text shows the results for the prediction of $pneu$, while Table \ref{tab:res_inf} shows these results for the prediction of $inf$. 

\begin{table}[t]
    \caption{Average precision over test set of three posterior probabilities for the diagnosis $inf$, each taking a different set of evidence into account (various combinations of background, symptoms and text). 
    We show mean ($\pm$ std) over 5 initialization seeds.}
    \label{tab:res_inf}
    \begin{tabularx}{\textwidth}{
    >{\hsize=0.72\hsize}X 
    >{\hsize=1.12\hsize\centering\arraybackslash}X 
    >{\hsize=1.08\hsize\centering\arraybackslash}X 
    >{\hsize=1.08\hsize\centering\arraybackslash}X 
    }
        \toprule
        & \multicolumn{3}{c}{\textbf{Average precision for $inf$}} \\
        \textbf{Model} & \textbf{$P(D_1\mid B,S_0,S_1,S_2,T)$} & \textbf{$P(D_1\mid B,S_0,S_1,S_2)$} & \textbf{$P(D_1\mid B,T)$} \\
        \hline
        \bn{} & - & 0.8884 ($\pm$ 0.0000) & - \\
        \bnplus{} & - & 0.9009 ($\pm$ 0.0000) & - \\
        \mbox{\ff{}} & 0.9042 ($\pm$ 0.0018) & 0.8813 ($\pm$ 0.0003) & 0.8821 ($\pm$ 0.0014) \\
        \mbox{\bngen{}} & 0.7968 ($\pm$ 0.0007) & 0.8889 ($\pm$ 0.0000) & 0.7624 ($\pm$ 0.0011) \\
        \mbox{\bndiscr{}} & 0.9016 ($\pm$ 0.0007) & 0.8889 ($\pm$ 0.0000) & 0.8738 ($\pm$ 0.0018) \\
        \bottomrule
    \end{tabularx}
\end{table}

\shrink
\subsection{Ablation Study} \label{sec:res_abl}
\shrink

In designing the DAG for models \bngen{} and \bndiscr{}, we explicitly included an arc between each diagnosis and text. This modeling decision makes sense when one assumes the presence of some unknown and unobserved symptoms in the text. 
In this section, we investigate how the models would perform if these relations were left out. 
We first introduce our generative and discriminative ablated models, and then discuss the empirical results. 

\shrink\shrink
\subsubsection{Ablated BN with Text Generator (\bngenabl{})} We remove the arcs $D_0 \rightarrow T$ and $D_1 \rightarrow T$ from the \bngen{} model shown in Figure \ref{fig:models}, forming the \bngenabl{} model. The text node now has only three parents (symptoms $S_0$, $S_1$ and $S_2$), meaning only 8 conditional Gaussians have to be fitted. Note that this means there are more text embeddings available to fit each Gaussian than there were for the \bngen{} model. The new joint distribution modeled by this Bayesian network is shown in Equation \ref{eq:joint_gen_abl}. Note that it differs from Equation \ref{eq:joint_gen} only in its definition of the conditional text distribution. We train this model in the same way as before, with the hyperparameters described in Section \ref{sec:app:training}.
\shrink
\begin{multline} \label{eq:joint_gen_abl}
\mathcal{P}(B,D_0,D_1,S_0,S_1,S_2,T) = \mathcal{P}(B)\mathcal{P}(D_0\mid B)\mathcal{P}(D_1\mid B)\mathcal{P}(S_0\mid D_0)\\
\mathcal{P}(S_1\mid D_0,D_1)\mathcal{P}(S_2\mid D_1)\mathcal{P}(T\mid S_0,S_1,S_2)
\end{multline}
\shrink\shrink\shrink

Bayesian inference over the ablated DAG partially differs from inference over the original DAG. The calculation of both $\mathcal{P}(D_i\mid B,S_0,S_1,S_2)$ and 
$\mathcal{P}(D_i\mid B,T)$ incurs only minimal changes (just swap out $\mathcal{P}(T\mid D_0,D_1,S_0,S_1,S_2)$ for $\mathcal{P}(T\mid S_0,S_1,S_2)$ in Equations \ref{eq:gen_inf_sympt} and \ref{eq:gen_inf_text}). However, the DAG shows that $D_i$ is independent of $T$ when all symptoms are known (no unblocked paths), meaning that $\mathcal{P}(D_i\mid B,S_0,S_1,S_2,T)$ = $\mathcal{P}(D_i\mid B,S_0,S_1,S_2)$.


\shrink
\subsubsection{Ablated BN with Text Discriminator (\bndiscrabl{})} We remove the arcs $T \rightarrow D_0$ and $T \rightarrow D_1$ from the \bndiscr{} model shown in Figure \ref{fig:models}, forming the \bndiscrabl{} model. This means that there are only 8 classifiers to be learned, as $\mathcal{P}(D_0\mid B)$ and $\mathcal{P}(D_1\mid B)$ can now be modeled as simple CPTs, 
just like $\mathcal{P}(B)$. The joint distribution for the ablated model is shown in Equation \ref{eq:joint_discr_abl}. We train this model with the hyperparameters described in Section \ref{sec:app:training}, using a learning rate of $0.05$ to learn the parameters of the CPTs for $D_0$, $D_1$ and $B$.

\shrink\shrink\shrink
\begin{multline} \label{eq:joint_discr_abl}
\mathcal{P}(B,D_0,D_1,S_0,S_1,S_2\mid T)
= \mathcal{P}(B)\mathcal{P}(D_0\mid B)\mathcal{P}(D_1\mid B)\mathcal{P}(S_0\mid D_0,T)\\
\mathcal{P}(S_1\mid D_0,D_1,T)\mathcal{P}(S_2\mid D_1,T)
\end{multline}
\shrink\shrink

Again, Bayesian inference over the ablated DAG partially differs from inference over the original DAG. $\mathcal{P}(D_i\mid B,S_0,S_1,S_2,T)$ is calculated analogously to Equation \ref{eq:inf_discr}, but with $\mathcal{P}(D_i\mid B)$ instead of $\mathcal{P}(D_i\mid B,T)$. 
Finally, it is clear from the ablated DAG that the diagnoses are independent of the text if no symptoms are observed (all paths between $D_i$ and $T$ are blocked by unobserved colliders). Therefore $\mathcal{P}(D_i\mid B,T)$ equals $\mathcal{P}(D_i\mid B)$, meaning the \bndiscrabl{} model cannot extract any information from the text without any observed symptoms.


\begin{table}[t]
    \caption{Average precision over test set for the ablated text models, which do not explicitly include the relation between diagnoses and text.}
    \label{tab:res_abl_pneu}
    \begin{tabularx}{\textwidth}{
    >{\hsize=0.72\hsize}X 
    >{\hsize=1.12\hsize\centering\arraybackslash}X 
    >{\hsize=1.08\hsize\centering\arraybackslash}X 
    >{\hsize=1.08\hsize\centering\arraybackslash}X
    }
        \toprule
        & \multicolumn{3}{c}{\textbf{Average precision for $pneu$}} \\
        \textbf{Model} & \textbf{$P(D_0\mid B,S_0,S_1,S_2,T)$} & \textbf{$P(D_0\mid B,S_0,S_1,S_2)$} & \textbf{$P(D_0\mid B,T)$} \\
        \hline
        \mbox{\bngenabl{}} & 0.0892 ($\pm$ 0.0007) & 0.0892 ($\pm$ 0.0007) & 0.0933 ($\pm$ 0.0009) \\
        \mbox{\bndiscrabl{}} & 0.1017 ($\pm$ 0.0008) & 0.1041 ($\pm$ 0.0072) & 0.0302 ($\pm$ 0.0000) \\
        \bottomrule
        \toprule
        & \multicolumn{3}{c}{\textbf{Average precision for $inf$}} \\
        \textbf{Model} & \textbf{$P(D_1\mid B,S_0,S_1,S_2,T)$} & \textbf{$P(D_1\mid B,S_0,S_1,S_2)$} & \textbf{$P(D_1\mid B,T)$} \\
        \hline
        \mbox{\bngenabl{}} & 0.8889 ($\pm$ 0.0000) & 0.8889 ($\pm$ 0.0000) & 0.8914 ($\pm$ 0.0002) \\
        \mbox{\bndiscrabl{}} & 0.8065 ($\pm$ 0.0004) & 0.8889 ($\pm$ 0.0000) & 0.4441 ($\pm$ 0.0000) \\
        \bottomrule
    \end{tabularx}
    \vspace{-40pt}
\end{table}


\shrink\shrink
\subsubsection{Analysis} 
Comparing the $pneu$ portion of Table \ref{tab:res_abl_pneu} with Table \ref{tab:res_pneu}, we immediately note that performance drops dramatically in the ablated versions of both the generative and discriminative model. While $\mathcal{P}(D_0\mid B,S_0,S_1,S_2)$ is still very similar to the \bn{} baseline, including text in the prediction $\mathcal{P}(D_0\mid B,S_0,S_1,S_2,T)$ now does not improve performance. Since we do not model the relation between diagnoses and text, the model can only extract information from the text \textit{through} the three symptoms we explicitly include in the DAG: $dysp$, $cough$ and $nasal$. Information regarding other useful symptoms, $pain$ and $fever$, cannot be extracted. 

While the \bngenabl{} model is able to extract the necessary information on the symptoms $S_0$, $S_1$ and $S_2$ from the text alone ($\mathcal{P}(D_0\mid B,T) \sim \mathcal{P}(D_0\mid B,S_0,S_1,S_2)$), \bndiscrabl{} performs abysmally when only text is included in the evidence. This comes as no surprise when we actually study the DAG: the diagnoses are independent of the text if no symptoms are observed. 
These independence assumptions do not match the reality we are trying to capture.


Comparing the $inf$ portion of Table \ref{tab:res_abl_pneu} with Table \ref{tab:res_inf} shows lower performance of \bndiscrabl{} compared to \bndiscr{} when only taking text into account ($\mathcal{P}(D_1\mid B,T)$). 
Conversely, \bngenabl{} actually improves over \bngen{} on both $\mathcal{P}(D_1\mid B,S_0,S_1,S_2,T)$ and $\mathcal{P}(D_1\mid B,T)$. Since the text node $T$ now only has three parents instead of five, there's more text embeddings available to fit each conditional Gaussian. 
Combined with the fact that there is no additional information in the text that can help to predict $inf$ anyway, modeling the direct relation between diagnosis and text will only result in a less reliable fit of the text distribution by the \bngen{} model.

\end{document}